\documentclass[letterpaper]{article} % DO NOT CHANGE THIS
\usepackage{aaai2026}  % DO NOT CHANGE THIS
\usepackage{times}  % DO NOT CHANGE THIS
\usepackage{helvet}  % DO NOT CHANGE THIS
\usepackage{courier}  % DO NOT CHANGE THIS
\usepackage[hyphens]{url}  % DO NOT CHANGE THIS
\usepackage{graphicx} % DO NOT CHANGE THIS
\usepackage{natbib}  % DO NOT CHANGE THIS AND DO NOT ADD ANY OPTIONS TO IT
\usepackage{caption} % DO NOT CHANGE THIS AND DO NOT ADD ANY OPTIONS TO IT
\usepackage{algorithm}
\usepackage{algorithmic}
\usepackage{amsfonts}
\usepackage{amsmath}
\usepackage{booktabs}
\usepackage{multirow}
\usepackage{subcaption}

\usepackage{newfloat}
\usepackage{listings}
\DeclareCaptionStyle{ruled}{labelfont=normalfont,labelsep=colon,strut=off} % DO NOT CHANGE THIS
\floatstyle{ruled}
\newfloat{listing}{tb}{lst}{}
\floatname{listing}{Listing}
\title{Dynamic Deep Graph Learning for Incomplete Multi-View Clustering with Masked Graph Reconstruction Loss}
\author {
    Zhenghao Zhang\textsuperscript{\rm 1},
    Jun Xie\textsuperscript{\rm 2},
    Xingchen Chen\textsuperscript{\rm 3},
    Tao Yu\textsuperscript{\rm 4},
    Hongzhu Yi\textsuperscript{\rm 1},
    Kaixin Xu\textsuperscript{\rm 5},
    Yuanxiang Wang\textsuperscript{\rm 1},
    Tianyu Zong\textsuperscript{\rm 1},
    Xinming Wang\textsuperscript{\rm 4},
    Jiahuan Chen\textsuperscript{\rm 6},
    Guoqing Chao\textsuperscript{\rm 7}\thanks{Corresponding authors.},
    Feng Chen\textsuperscript{\rm 2},
    Zhepeng Wang\textsuperscript{\rm 2},
    Jungang Xu\textsuperscript{\rm 1}\footnotemark[1]
}
\affiliations {
    \textsuperscript{\rm 1}School of Computer Science and Technology, University of Chinese Academy of Sciences, Beijing\\
    \textsuperscript{\rm 2}Lenovo Research, Beijing\\
    \textsuperscript{\rm 3}Faculty of Computing, Harbin Institute of Technology, Harbin\\
    \textsuperscript{\rm 4}Institute of Automation, Chinese Academy of Sciences, Beijing\\
    \textsuperscript{\rm 5}School of Software Technology, Zhejiang University, Ningbo\\
    \textsuperscript{\rm 6}School of Automation and Intelligent Sensing, Shanghai Jiao Tong University, Shanghai\\
    \textsuperscript{\rm 7}School of Computer Science and Technology, Harbin Institute of Technology, Weihai\\
    zhangzhenghao25@mails.ucas.ac.cn, guoqingchao10@gmail.com, xujg@ucas.ac.cn\\ 
}

\begin{document}
\maketitle
\begin{abstract}
The prevalence of real-world multi-view data makes incomplete multi-view clustering (IMVC) a crucial research. The rapid development of Graph Neural Networks (GNNs) has established them as one of the mainstream approaches for multi-view clustering. Despite significant progress in GNNs-based IMVC, some challenges remain: (1) Most methods rely on the K-Nearest Neighbors (KNN) algorithm to construct static graphs from raw data, which introduces noise and diminishes the robustness of the graph topology. (2) Existing methods typically utilize the Mean Squared Error (MSE) loss between the reconstructed graph and the sparse adjacency graph directly as the graph reconstruction loss, leading to substantial gradient noise during optimization. To address these issues, we propose a novel \textbf{D}ynamic Deep \textbf{G}raph Learning for \textbf{I}ncomplete \textbf{M}ulti-\textbf{V}iew \textbf{C}lustering with \textbf{M}asked Graph Reconstruction Loss (DGIMVCM). Firstly, we construct a missing-robust global graph from the raw data. A graph convolutional embedding layer is then designed to extract primary features and refined dynamic view-specific graph structures, leveraging the global graph for imputation of missing views. This process is complemented by graph structure contrastive learning, which identifies consistency among view-specific graph structures. Secondly, a graph self-attention encoder is introduced to extract high-level representations based on the imputed primary features and view-specific graphs, and is optimized with a masked graph reconstruction loss to mitigate gradient noise during optimization. Finally, a clustering module is constructed and optimized through a pseudo-label self-supervised training mechanism. Extensive experiments on multiple datasets validate the effectiveness and superiority of DGIMVCM.
\end{abstract}

\begin{links}
    \link{Code}{https://github.com/PaddiHunter/DGIMVCM}
    % \link{Datasets}{https://aaai.org/example/datasets}
    % \link{Extended version}{https://aaai.org/example/extended-version}
\end{links}

\section{Introduction}
Multi-view data refers to data that describes the same object from multiple perspectives~\citep{fu2020overview}. Multi-view learning aims to leverage information from multiple views to enhance generalization performance~\citep{yu_review_2024,xu2013survey}. Multi-view clustering (MVC) is a prominent unsupervised learning branch within multi-view learning, which groups data by exploiting the consistency and complementarity across views \citep{fang2023comprehensive,yang2018multi,zhou2024survey, wu2024self}. Most existing multi-view clustering methods~\citep{xu2021deep} operate under a data completeness assumption. However, due to various unavoidable factors, samples often have only partial views available~\citep{tang2024incomplete, yu2025incomplete}. Consequently, incomplete multi-view clustering has garnered increasing attention.

Numerous studies~\citep{renmulti, ren2024dynamic} leverage Graph Neural Networks (GNNs) to address multi-view clustering problems. Most of these methods construct adjacency graphs from raw data using the K-nearest neighbors algorithm, and these graphs remain fixed during model training \citep{chao2025global,chao2024incomplete}. However, such pre-defined graphs often suffer from noise inherent in the raw data, and their immutability during training can limit performance improvements. To mitigate this, some research endeavors first denoise the data using autoencoders and subsequently extract graph structures from the refined features \citep{huang2023self, wang2024surer}. Nevertheless, these approaches are not well-suited to effectively handle data incompleteness. Furthermore, many methods \citep{chaofederated, wang2024surer,cheng2021multi} directly compute the MSE loss between the reconstructed graph and the adjacency graph as graph reconstruction loss. Due to the sparsity of adjacency graphs, this often leads to the loss optimization gradient direction containing substantial noise, which impedes further performance enhancement.

To address these limitations, we propose the Dynamic Deep Graph Learning for Incomplete Multi-View Clustering with Masked Graph Reconstruction Loss. Our approach first utilizes a Graph Convolutional Network(GCN)-based embedding layer with a global graph to impute features for missing views. Subsequently, view-specific graphs are dynamically extracted from these denoised features. The structures of these view-specific graphs are then refined by global graph, and consistency among them is learned through graph structure contrastive learning. We then design a Graph Attention Network (GAT) encoder, enabling the model to dynamically learn edge weights, and optimize it using the masked graph reconstruction loss to reduce gradient noise during optimization. Finally, a clustering module is integrated, leveraging pseudo-labels to guide the training process.

The main contributions of this paper are summarized as follows:
\begin{enumerate}
    \item We propose a novel incomplete multi-view clustering method that dynamically learns graph structures and edge weights. Extensive experimental results on four widely used datasets validate the superiority of it.
    \item We introduce an innovative embedding layer and a graph structure contrastive learning method, which are designed to facilitate robust adjacency graph extraction, enable the learning of inter-view consistent graph structures, and effectively handle data incompleteness.
    \item We present a generic masked graph reconstruction loss that effectively mitigates gradient noise during the optimization process. This is demonstrated through both theoretical analysis and extensive experimental evaluation.
\end{enumerate}

\section{Related Works}
\subsection{Deep Incomplete Multi-view Clustering}
Deep incomplete multi-view clustering (DIMVC) methods are broadly categorized into imputation-based and imputation-free approaches. Imputation-based methods typically initiate by recovering missing views, followed by clustering on the reconstructed complete multi-view dataset. For example, AGDIMC \citep{pu2024adaptive} imputes missing features adaptively by leveraging cross-view soft cluster assignments and global cluster centroids. Similarly, \citet{tang2022deep} propose a bi-level optimization framework that dynamically imputes missing views from learned semantic neighbors. Conversely, imputation-free methods directly perform clustering based on available views. For instance, \citet{xu2024deep} utilizes multiple view-specific encoders to extract information from each view and employs the Product-of-Experts (PoE) approach to obtain a common representation.

\subsection{Graph Neural Networks for Multi-view Clustering}
Graph Neural Networks (GNNs) are widely applied in multi-view clustering due to their ability to learn both node attributes and graph structures~\citep{wen2021structural,shao2022heterogeneous,wen2024dual}. For instance, SERIES~\citep{wang2024contrastive} leverages GNNs through deep graph autoencoders to extract latent representations from multi-view data, which are then fused to form a consistent structural graph for clustering. SURER~\citep{wang2024surer} enhances multi-view clustering by adaptively refining view-specific graphs and unifying them into a heterogeneous graph, subsequently learning a consensus representation via graph neural network. SGDMC~\citep{huang2023self} performs multi-view clustering using self-supervised graph attention networks, where attention allocation considers both node attributes and pseudo-labels. GHICMC~\citep{chao2025global} performs multi-view clustering by learning view-specific and consensus representations via GCNs, imputing missing data through hierarchical global graph propagation. ICMVC~\citep{chao2024incomplete} leverages multi-view consistency transfer with GCN for missing data, fuses view-specific representations via instance-level attention, and applies contrastive learning and high-confidence guiding.

\section{The Proposed Method}
\begin{figure*}[ht]
	\centering
	\includegraphics[width=0.9\textwidth]{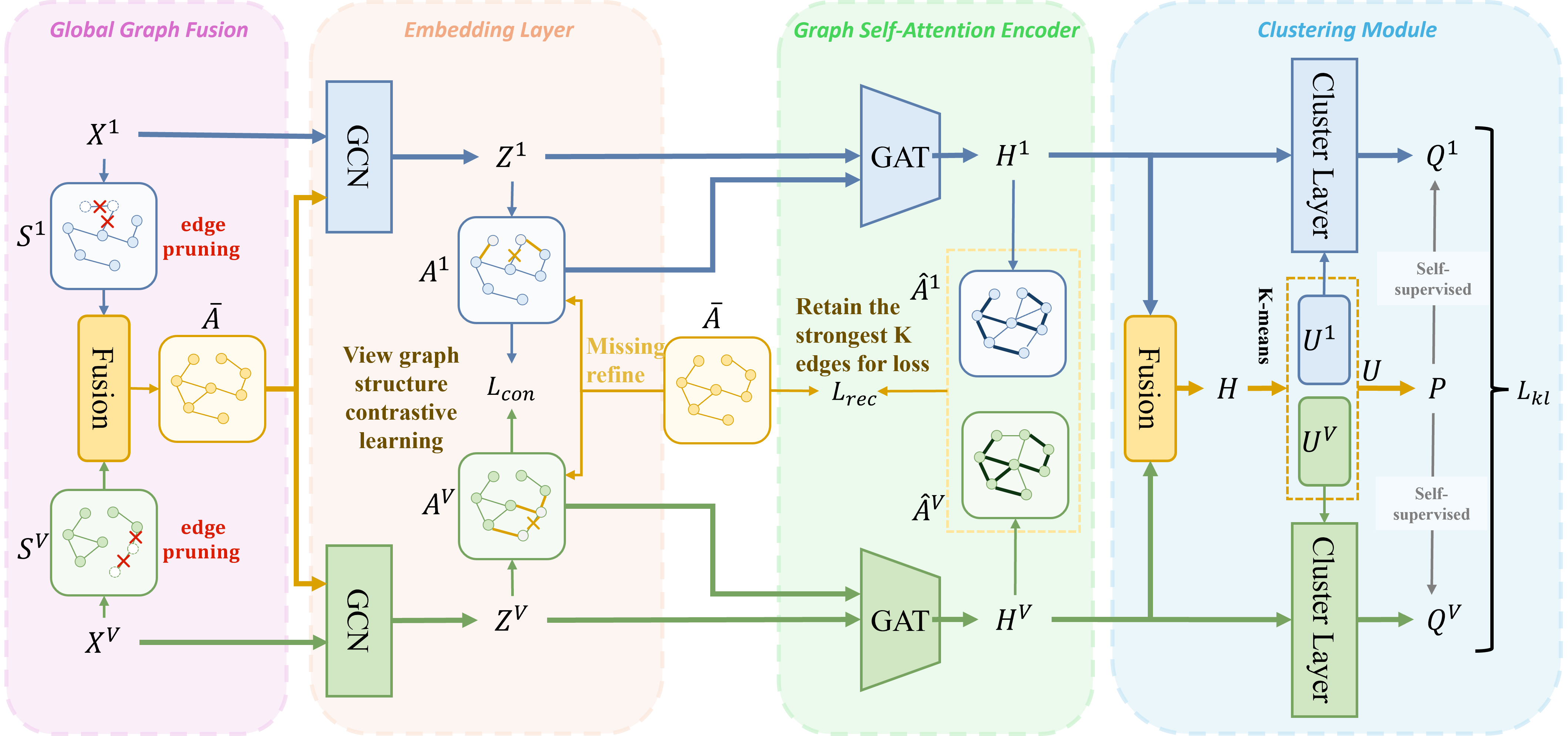}
	\caption{The overview of the DGIMVCM framework. The framework comprises four components: global graph fusion, embedding layer, graph self-attention encoder, and the clustering module. Initially, a global graph $\overline{A}$ is constructed by fusing view-specific similarity matrices with missing sample edge pruning. Subsequently, a GCN-based embedding layer utilizes $\overline{A}$ to extract imputed primary features $\{Z^v\}_{v=1}^V$ and view-specific graphs $\{A^v\}_{v=1}^V$ with missing structure refinement by the global graph, which is optimized via view-specific graph structure consistency contrastive learning. Next, a graph self-attention encoder is employed to extract high-level features $\{H^v\}_{v=1}^V$, optimized by a masked graph reconstruction loss that focuses only on the $K$ strongest edges. Finally, a clustering module performs self-supervised training using pseudo-labels and obtain the final clustering results.}
	\label{fig:overview}
\end{figure*}

In this section, we introduce the proposed method consisting of four modules: global graph fusion, embedding layer, graph self-attention encoder and the clustering module. Figure~\ref{fig:overview} illustrates the overall architecture of the method.
\subsubsection{Notations}
Let $\mathcal{X}=\{\mathcal{X}^1,...,\mathcal{X}^v,...,\mathcal{X}^V\}$ denotes a complete multi-view dataset with $V$ views. $\mathcal{X}^v \in \mathbb{R}^{N\times d^v}$ represents the dataset for the $v$-th view, where $N$ is the number of samples and $d^v$ is the feature dimension of the $v$-th view. To construct an incomplete multi-view dataset $X$, a missing indicator matrix $M\in\{0,1\}^{N\times V}$ is defined. Specifically, $M_{iv}=1$ indicates the presence of the $v$-th view for the $i$-th sample, while $M_{iv}=0$ denotes the absence of it. Each sample has at least one available view. The incomplete dataset is then defined as $X=\{X^1,...,X^v,...X^V\}$, where $X^v = \mathcal{M}^v\odot\mathcal{X}^v$. Here, $\mathcal{M}\in\{0,1\}^{N\times d^v}$ is a matrix obtained by broadcasting the $v$-th column of $M$, and $\odot$ denotes the Hadamard product. The objective of the algorithm is to cluster $X$ into $K$ categories.

\subsection{Global Graph Fusion with Missing Robustness}
To exploit consistency across views and leverage complementary information from other views to alleviate missing data issues, we construct a global graph for feature imputation in the embedding layer and optimization of the encoder.

The global graph is aggregated from the graph structures of individual views. Specifically, we first compute the similarity matrix $S^v \in (0,1]^{N\times N}$ for each view using the radial basis function: $S_{ij}^v = \exp\left(-\frac{\|x_i^v - x_j^v\|^2_2}{t}\right)$,
where $S_{ij}^v$ quantifies the similarity between samples $x^v_i$ and $x^v_j$ in the $v$-th view, and $t$ is a scale parameter. A challenge arises because missing samples are represented by zero vectors, leading to abnormally high similarities between them, which significantly impact the edges of the global graph during aggregation. Therefore, prior to aggregation, each $S^v$ is pre-processed by pruning edges associated with missing samples. The global graph $\overline{A}\in\{0,1\}^{N\times N}$ is then computed as follows:
\begin{equation}
    \overline{A}=\text{TopK}\left(\sum_{v=1}^Vf_P(S^v)\right),
    \label{eq:global_graph}
\end{equation}
where the $f_P(\cdot)$ function re-calibrates the similarity matrix by zeroing out rows and columns associated with missing samples. Subsequently, the function $\text{TopK}(\cdot)$ constructs an adjacency matrix by setting the $K$ largest elements in each row to 1 and the remainder to 0. This approach effectively reduces the influence of missing samples on the global graph.

\subsection{Embedding Layer for Dynamic View-specific Graph Construction}
Current GNN-based methods for IMVC are often limited by static graph structure with raw data noise. To overcome these challenges, we introduces a Graph Convolutional Network (GCN)-based embedding layer to dynamically update the refined view-specific graph structure and imputes the feature of missing samples. Taking view $v$ as an example, the embedding layer is designed as follows.

\subsubsection{Architecture of the Embedding Layer}
We implement the embedding layer using GCN with the global graph as input. This approach allows the primary features of missing samples to integrate information from their adjacent complete samples. The computation for the embedding layer of view $v$ is given by:
\begin{equation}
    Z^v=\overline{A}'X^vW_e^v+b_e^v,
    \label{eq:embed_layer}
\end{equation}
where $Z^v$ represents the primary features for the $v$-th view, $\overline{A}'=\overline{D}^{-\frac{1}{2}}\overline{A}\space\overline{D}^{-\frac{1}{2}}$ is the normalized global graph adjacency matrix, and $\overline{D}$ is a diagonal matrix with $\overline{D}_{ii}=\sum_j\overline{A}_{ij}$. Furthermore, $W_e^v$ and $b_e^v$ are the trainable parameters in the embedding layer.

\subsubsection{Dynamic View-specific Graph Construction}
Similar to the global graph construction, we first compute a similarity matrix $\hat{S}^v \in (0,1]^{N \times N}$ based on $Z^v$ by the radial basis function as previously described. At this stage, the similarity matrix is computed using the imputed features and integrates information from the global graph, thereby possessing enhanced representational power. Subsequently, we construct the view-specific adjacency matrix $A^v \in \{0,1\}^{N \times N}$ using the $\text{TopK}(\cdot)$ function:
\begin{equation}
    A^v = \text{TopK}(\hat{S}^v),
    \label{eq:view_graph}
\end{equation}
where the function $\text{TopK}(\cdot)$ is defined consistently with its usage in Eq.~\eqref{eq:global_graph}. To mitigate the impact of missing data, following related works~\citep{chaofederated}, $\mathbf{A}^v$ undergoes a refinement process: inaccurate edges corresponding to missing samples are replaced with their counterparts from $\overline{\mathbf{A}}$.

\subsubsection{Contrastive Learning for View-specific Graphs}
Direct application of contrastive learning to features can force multi-view features into a single representation space, thereby reducing the diversity of feature representation. To circumvent this issue, we instead employ view-graph structure contrastive learning to simultaneously uncover consistency in graph structures across different views and enhance the distinctiveness of graph structures among samples. Specifically, graph structures of the same sample across different views are considered positive pairs, whereas graph structures of different samples are considered as negative pairs. The contrastive loss for graph structures between view $v$ and view $w$ is then calculated as:
\begin{equation}
    {L}^{(v,w)}_{con}=\frac{1}{N}\sum_{i=1}^N-\log{\frac{\exp(d(\hat{s}^v_i,\hat{s}^w_i)/\tau)}{\sum_{u=v,w}\sum_{j\neq i}\exp(d(\hat{s}^v_i,\hat{s}^u_j)/\tau)}},
    \label{eq:loss_con_v}
\end{equation}
where $d(\hat{s}^v_i,\hat{s}^w_i)$ denotes the cosine similarity between $\hat{s}^v_i$ and $\hat{s}^w_i$, $\hat{s}^v_i$ represents the $i$-th row of the similarity matrix $\hat{S}^v$, and $\tau$ is the temperature parameter. The contrastive loss for graph structures across different views is given by:
\begin{equation}
    L_{con}=\sum_{v=1}^V\sum_{w\neq v}L_{con}^{(v,w)}.
    \label{eq:loss_con}
\end{equation}

\subsection{Graph Self-Attention Encoder for Representation Learning}

\subsubsection{Architecture of the Encoder}
To overcome the limitations of fixed edge weights inherent in GCN, we employ Graph Attention Networks (GATs) to construct the encoder. GATs can automatically learn self-attention weights, which serve as dynamic edge weights. Specifically, for view $v$, the encoder takes $Z^v$ and $A^v$ as input, mapping them to refined high-level feature representations $H^v \in \mathbb{R}^{N \times \hat{d}^v}$, where $\hat{d}^v$ is the dimension of high-level feature for view $v$.

Considering the $l$-th layer, the core of the encoder involves computing the self-attention weight $e^v_{ij,(l)}$ between sample $i$ and sample $j$. This is achieved using a Dot-Product Attention mechanism~\citep{vaswani2017attention}:
\begin{equation}
   e^v_{ij,(l)} = \left(H^v_{i,(l-1)} W^v_{Q,(l)}\right) \left(H^v_{j,(l-1)} W^v_{K,(l)}\right)^\mathrm{T},
   \label{eq:e}
\end{equation}
where $H^v_{i,(l-1)}$ denotes the feature representation of sample $i$ in view $v$ at the $(l-1)$-th layer of the encoder, $W^v_{Q,(l)}$ and $W^v_{K,(l)}$ are the learnable projection matrices for computing the Query and Key respectively at the $l$-th layer. This enables the model to adaptively learn attention weights. Notably, the features for the 0-th layer are initialized as $H^v_{(0)}=Z^v$.

Next, to integrate the original graph structure information, we constrain the computed attention weight matrix using the view-specific graph and normalize it row-wise using softmax function, yielding the edge weight matrix $\tilde{A}^v_{(l)}$:
\begin{equation}
   \tilde{A}_{ij,(l)}^v = \begin{cases}
       \frac{\exp({e^v_{ij,(l)}})}{\sum\limits_{k \in \{k \mid A^v_{ik}=1\}} \exp({e^v_{ik,(l)}})}, & \text{if } A^v_{ij}=1 \\
       0, & \text{if } A^v_{ij}=0,
   \end{cases}
   \label{eq:attention}
\end{equation}
where $A^v_{ij}=1$ indicates the existence of an edge between samples $i$ and $j$ in the original view-specific graph $A^v$. This operation ensures that self-attention edge weights exist only between originally connected nodes.

Utilizing this learnable edge weight matrix $\tilde{A}^v_{(l)}$, the features $H^v_{(l)}$ at the $l$-th layer are computed as follows:
\begin{equation}
   H^v_{(l)}=\sigma((\tilde{A}^v_{(l)}H^v_{(l-1)}{W^v_{V,(l)}})W^v_{(l)}+b^v_{(l)})+H^v_{(l-1)},
   \label{eq:encoder}
\end{equation}
where $W^v_{V,(l)}$ is a projection matrix for computing the Value, $b^v_{(l)}$ is the bias vector, $\sigma(\cdot)$ denotes the activation function, and $W^v_{(l)}$ is a projection matrix used for further feature transformation. It is noteworthy that residual connections are employed to mitigate degradation issues in networks.

\subsubsection{Masked Graph Reconstruction Loss}
For simplicity, we compute a similarity matrix serving as the reconstructed graph $\hat{A}^v$ using the radial basis function:
\begin{equation}
   \hat{A}^v_{ij}=e^{-\frac{||h_i^v-h_j^v||^2_2}{t}},
   \label{eq:rec_graph}
\end{equation}
where $h_i^v$ represents the high-level feature of the $i$-th sample in view $v$. To guide the features from different views towards learning a unified graph structure, the global graph $\overline{A}$ serves as the reconstruction target. Unlike conventional graph reconstruction losses, the masked graph reconstruction loss focuses on the strongest $k$ edges for each sample within the reconstructed graph, and is computed as:
\begin{equation}
   L_{rec}=\frac{1}{V}\sum_{v=1}^V\frac{1}{N}||M^v\odot\hat{A}^v-\overline{A}||^2_2,
   \label{eq:loss_rec}
\end{equation}
where $M^v\in\{0,1\}^{N\times N}$ is the graph mask matrix. Let $\hat{A}^v_{i\cdot}$ denote the $i$-th row of $\hat{A}^v$, then $M^v_{ij}=1$ if $\hat{A}^v_{ij}$ is among the top-$k$ largest elements in $\hat{A}^v_{i\cdot}$, otherwise $M^v_{ij}=0$.

\subsubsection{Gradient Analysis of Masked Graph Reconstruction Loss}
This subsection highlights the advantages of our masked graph reconstruction loss over traditional approaches by analyzing their gradients. Detailed derivations are provided in Appendix A. Here, we present only the results.

Without loss of generality, consider the loss associated with $i$-th sample feature $h^v_i$ from view $v$. We first analyze the gradient of the traditional graph reconstruction loss $L_{rec\_t,i}^v=\sum_{j=1}^N(\hat{A}^v_{ij}-\overline{A}_{ij})^2$, with respect to $h^v_i$:
\begin{align}
   \frac{\partial L^v_{rec\_t,i}}{\partial h^v_i}=\frac{4}{t}(&\sum_{j\in E_i}(\hat{A}^v_{ij}-1)\hat{A}^v_{ij}(h^v_j-h^v_i)+ \notag\\
   &\sum_{j\notin E_i}{(\hat{A}^v_{ij})}^2(h^v_j-h_i^v)),
   \label{eq:loss_rec_t}
\end{align}
where $E_i=\{j \mid \overline{A}_{ij}=1\}$ denotes the set of indices for samples adjacent to sample $i$ in the global graph. Two types of sample loss gradient are considered. For $j\in E_i$, the term $(\hat{A}^v_{ij}-1)\hat{A}^v_{ij}$ is negative. Consequently, the gradient direction is $h^v_i-h^v_j$. As the model updates in the direction opposite to the loss gradient, this implies that $h_i^v$ is updated towards $h_j^v$, which can be interpreted as an attractive force. Conversely, if $j\notin E_i$, $h_i^v$ is updated away from $h_j^v$, acting as a repulsive force. Attraction forces are considered more critical than repulsive forces, as they facilitate the formation of clusters. However, because $k$ is generally small, leading to $|E_i| \ll N$, the second term in Eq.~\eqref{eq:loss_rec_t} encompasses a large number of components. Moreover, in the early stages of training, $\hat{A}^v_{ij}$ values are not negligible, and the repulsive forces exhibit diverse directions. This introduces significant gradient noise during optimization. Additionally, as the global graph is sparse, non-adjacent samples may still belong to the same cluster. The repulsive forces between such samples can excessively reduce their similarity.

For clarity, we refer to edges present in the global graph as ``informative edges''. Edges that are absent in the global graph but present in the reconstructed graph are termed ``hard edges'', as they indicate samples that are difficult for the encoder to distinguish. The gradient of our masked graph reconstruction loss $L_{rec\_m,i}^v=\sum_{j=1}^N(M^v_{ij}\hat{A}^v_{ij}-\overline{A}_{ij})^2$, with respect to $h_i^v$ is given by:
\begin{align}
    \frac{\partial L^v_{rec\_m,i}}{\partial h^v_i}
    &=\frac{4}{t}(\sum_{j:j\in E_i,M^v_{ij}=1}(\hat{A_{ij}^v}-1)\hat{A}^v_{ij}(h^v_j-h^v_i) \notag\\
    &+\sum_{j:j\notin E_i,M^v_{ij}=1}{\hat{A_{ij}^v}}^2(h^v_j-h^v_i)).
\end{align}
Here, the first term represents informative edges, while the second term addresses hard edges. During optimization, the masking mechanism selects and retains the $k$ strongest edges in the reconstructed graph. The small value of $k$ significantly reduces gradient noise, predominantly manifesting as attractive forces, thereby enabling $h^v_i$ to be optimized towards well-defined directions. For the selected edges, if they are also present in the global graph, the loss reinforces these connections. If they are absent, the loss weakens these connections, allowing other more relevant edges to be selected while preventing excessive weakening.

\begin{algorithm}
	\caption{Optimization algorithm for DGIMVCM}
        \textbf{Input:} Incompleted multi-view data with $N$ samples and $V$ views $\{X^v\}_{v=1}^V$, number of clusters $K$, maximum iterations $epochs$.\\
	\textbf{Output:} Clustering results $Y$.
	\begin{algorithmic}[1]

            \STATE Compute global graph $\overline{A}$ by Eq.~\eqref{eq:global_graph}

            \FOR{epoch=1 to $epochs$}
            \STATE Compute primary features $\{Z^v\}_{v=1}^V$ by Eq.~\eqref{eq:embed_layer}
            \STATE Compute view-specific graphs ${\{A^v\}}_{v=1}^V$ by Eq.~\eqref{eq:view_graph}
            \STATE Compute the loss $L_{con}$ by Eq.~\eqref{eq:loss_con}
            \STATE Compute the high-level features $\{H^v\}_{v=1}^V$ by Eq.~\eqref{eq:encoder}
            \STATE Compute the $\{\hat{A}^v\}_{v=1}^V$ by Eq.~\eqref{eq:rec_graph}
            \STATE Compute the loss $L_{rec}$ by Eq.~\eqref{eq:loss_rec}
            \STATE Compute the fused features $H$ by Eq.~\eqref{eq:feature_fusion}
            \STATE Perform K-means on $H$ to obtain $U$
            \STATE Compute $P$ by Eq.~\eqref{eq:p}
            \STATE Compute $\{Q^v\}_{v=1}^V$ by Eq.~\eqref{eq:q_v}
            \STATE Compute the loss $L_{kl}$ by Eq.~\eqref{eq:loss_kl}
            \STATE Compute the overall loss $L$ by Eq.~\eqref{eq:loss}
            \STATE Update through gradient descent to minimize $L$
            \ENDFOR
            \STATE Obtain the final clustering result $Y$ by Eq.~\eqref{eq:y}
	\end{algorithmic}
	\label{alg:DGIMVCM}
\end{algorithm}

\subsection{Self-supervised Clustering Module}

\subsubsection{Pseudo-label Acquisition}
To provide global clustering supervision, we first compute global pseudo-labels $P$. Specifically, High-level features $\{H^v\}_{v=1}^V$ from all views are fused to construct a global feature representation $H \in \mathbb{R}^{N \times \sum_{v=1}^V\hat{d}_v}$:
\begin{equation}
   H=[H^1, \dots, H^v, \dots, H^V],
   \label{eq:feature_fusion}
\end{equation}
where $[\cdot, \dots, \cdot]$ denotes the horizontal concatenation of matrices. Subsequently, during training, the K-means algorithm is applied to $H$ to iteratively update the global cluster centers $U=[U^1, \dots, U^V] \in \mathbb{R}^{K \times \sum_{v=1}^V\hat{d}_v}$. Here, $U^v \in \mathbb{R}^{K \times \hat{d}^v}$ represents the cluster centers for view $v$. Consequently, the soft labels $Q$ are computed as follows:
\begin{equation}
   q_{ij}=\frac{(1+||h_i-U_j||^2_2)^{-1}}{\sum_{k=1}^K(1+||h_i-U_k||^2_2)^{-1}},
   \label{eq:q}
\end{equation}
where $q_{ij}$ denotes the probability of the $i$-th sample being assigned to the $j$-th cluster, $h_i$ represents the global feature of the $i$-th sample, and $U_j$ is the $j$-th global cluster center. Finally, by sharpening the soft labels, the pseudo-labels $P$ are obtained via:
\begin{equation}
   p_{ij}=\frac{(\frac{q_{ij}}{\sum_j q_{ij}})^2}{\sum_j(\frac{q_{ij}}{\sum_j q_{ij}})^2}.
   \label{eq:p}
\end{equation}

\subsubsection{Clustering Layer}
The clustering layer computes soft labels based on the cluster centers $\{U^v\}_{v=1}^V$ using Student's t-distribution, as presented below:
\begin{equation}
   q^v_{ij}=\frac{(1+||h_i-U^v_j||^2_2)^{-1}}{\sum_{k=1}^K(1+||h_i-U_k^v||^2_2)^{-1}},
   \label{eq:q_v}
\end{equation}
where $q^v_{ij}$ denotes the probability that the $i$-th sample in view $v$ is assigned to the $j$-th cluster, $h^v_i$ denotes the high level feature of the $i$-th sample in view $v$, and $U^v_j$ denotes the $j$-th cluster center for view $v$, which implicitly updates with the global cluster centers $U$. The global pseudo-labels $P$ serve as the supervisory signal to optimize the soft distributions of each view. This is achieved by introducing kl divergence loss:
\begin{equation}
   L_{kl}=\sum_{v=1}^V D_{KL}(P||Q^v)=\sum_{v=1}^V\sum_{i=1}^N\sum_{j=1}^Kp_{ij}\log\frac{p_{ij}}{q^v_{ij}}.
   \label{eq:loss_kl}
\end{equation}
Finally, by combining the results from all view-specific soft labels, the ultimate clustering assignment $Y=\{y_i\}_{i=1}^N$ is obtained:
\begin{equation}
   y_i=\mathop{\arg \max}\limits_k\ \sum_{v=1}^Vq^v_{ik},
   \label{eq:y}
\end{equation}
where $y_i$ represents the assigned cluster label for the $i$-th sample.

\subsection{The Overall Loss Function}
In summary, the overall loss function is defined as:
\begin{equation}
   L=L_{rec}+\alpha L_{con}+\beta L_{kl},
   \label{eq:loss}
\end{equation}
where, $\alpha$ and $\beta$ represent hyperparameters that balance the contributions of the three loss functions. The complete  algorithn procedure is detailed in Algorithm \ref{alg:DGIMVCM}.

\section{Experiments}
\begin{table*}[ht]
\caption{\textbf{Experimental results on the four datasets. The best results in each column are shown in bold and the second best results are underlined.  $\delta=0$ indicates complete, while $\delta=0.5$ indicates incomplete.}}
\label{tab:baselines}
\centering
\begin{tabular}{cccccccccccccc}
\toprule
\multirow{2}{*}{$\delta$}&\multirow{2}{*}{Methods}&\multicolumn{3}{c}{HW}&\multicolumn{3}{c}{Scene-15} & \multicolumn{3}{c}{Landuse-21}&\multicolumn{3}{c}{100Leaves} \\
\cline{3-5}\cline{6-8}\cline{9-11}\cline{12-14}
~&~&ACC&NMI&ARI&ACC&NMI&ARI&ACC&NMI&ARI&ACC&NMI&ARI\\
\midrule
\multirow{8}{*}{0}
&DIMVC&0.446&0.533&0.381&0.350&0.309&0.179&0.243&0.301&0.109&0.825&0.922&0.753\\
&DSIMVC&0.759&0.756&0.661&0.281&0.299&0.146&0.177&0.173&0.049&0.401&0.725&0.294\\
&GIGA&0.807&0.853&0.756&0.221&0.263&0.041&0.131&0.257&0.017&0.742&0.877&0.483\\
&GIMVC&\underline{0.935}&0.886&\underline{0.874}&0.426&\underline{0.465}&\underline{0.279}&0.258&\underline{0.337}&0.114&0.857&0.952&0.819\\
&CDIMC-net&0.861&\underline{0.890}&0.827&0.347&0.421&0.198&0.184&0.236&0.054&0.799&0.938&0.751\\
&MRL\_CAL&0.478&0.526&0.337&0.194&0.168&0.069&0.163&0.169&0.045&0.224&0.587&0.126\\
&DCP&0.828&0.852&0.771&0.401&0.436&0.240&0.260&0.311&0.121&0.606&0.843&0.496\\
&GHICMC&0.854&0.860&0.794&\underline{0.434}&0.438&0.273&\underline{0.266}&0.313&\underline{0.128}&\underline{0.940}&\underline{0.969}&\underline{0.914}\\
&Ours&\textbf{0.979}&\textbf{0.953}&\textbf{0.953}&\textbf{0.503}&\textbf{0.508}&\textbf{0.330}&\textbf{0.319}&\textbf{0.393}&\textbf{0.171}&\textbf{0.956}&\textbf{0.984}&\textbf{0.940}\\
\midrule
\multirow{8}{*}{0.5}&DIMVC&0.322&0.255&0.151&0.310&0.261&0.143&0.226&0.278&0.099&0.579&0.731&0.380\\
&DSIMVC&0.729&0.687&0.586&0.260&0.267&0.125&0.172&0.169&0.048&0.295&0.616&0.171\\
&GIGA&0.764&0.730&0.594&0.146&0.127&0.008&0.182&0.279&0.025&0.418&0.649&0.055\\
&GIMVC&\underline{0.911}&0.838&\underline{0.825}&0.385&0.373&0.218&0.228&0.273&0.085&0.688&\underline{0.842}&0.555\\
&CDIMC-net&0.858&\underline{0.861}&0.792&0.217&0.268&0.067&0.122&0.161&0.020&0.330&0.643&0.207\\
&MRL\_CAL&0.358&0.370&0.191&0.189&0.150&0.065&0.162&0.168&0.044&0.145&0.434&0.052\\
&DCP&0.628&0.671&0.463&0.397&0.408&0.232&\underline{0.256}&\underline{0.292}&\underline{0.119}&0.449&0.689&0.209\\
&GHICMC&0.844 &0.844 &0.784 &\underline{0.413} &\underline{0.403} &\underline{0.247} &0.251 &0.285 &0.112 &\underline{0.707} &0.826 &\underline{0.565} \\
&Ours&\textbf{0.939}&\textbf{0.879}&\textbf{0.870}&\textbf{0.452}&\textbf{0.430}&\textbf{0.273}&\textbf{0.287}&\textbf{0.328}&\textbf{0.140}&\textbf{0.812}&\textbf{0.891}&\textbf{0.703}\\
\bottomrule
\end{tabular}
\end{table*}

\begin{figure*}[ht]
	\centering
	\includegraphics[width=\textwidth]{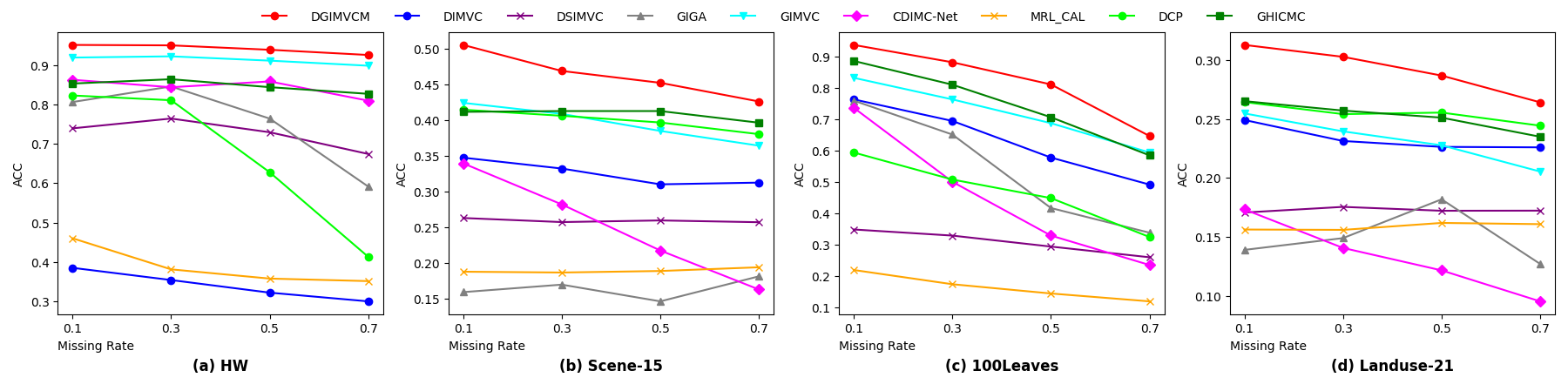}
        \vspace{-9mm}
	\caption{Accuracy on four datasets with different missing rates.}
	\label{fig:baselines}
\end{figure*}

\subsection{Experimental Settings}

To construct incomplete datasets, the missing rate is defined as $\delta=(N-n)/N$, which represents the proportion of samples with missing views in a multi-view dataset. Here, $N$ denotes the total number of samples in the dataset, and $n$ represents the number of samples with complete views. For incomplete samples, a random number of views are removed.

\subsection{Datasets and Metrics}
% The empirical evaluation of the proposed method is conducted on four commonly used multi-view datasets: Scene-15~\cite{fei2005bayesian}, HandWritten (HW)~\citep{li2015large}, Landuse-21~\citep{yang2010bag}, and 100leaves~\citep{mallah2013plant}. Statistical details for these datasets are presented in Table \ref{tab:datasets}.
Experiments are conducted on four widely used multi-view datasets. Specifically, \textbf{Scene-15}~\cite{fei2005bayesian,lazebnik2006beyond} comprises 4485 scene images categorized into 15 classes, with each sample represented by three distinct views. \textbf{HandWritten}~\citep{li2015large} contains 2000 samples across ten numeric categories, each characterized by six views. \textbf{Landuse-21}~\citep{yang2010bag} consists of 2100 satellite images distributed among 21 categories, with each category containing 100 images, and each image represented by three views. Finally, the \textbf{100leaves} dataset~\citep{mallah2013plant} includes 1600 image samples derived from 100 plant species, with each sample possessing three different views.

To evaluate the effectiveness of the clustering approach, three widely recognized metrics are employed: clustering accuracy (ACC), normalized mutual information (NMI), and adjusted random index (ARI). For each of these metrics, a higher value signifies superior clustering performance.

\subsection{Compared Methods}
To demonstrate the superiority of our proposed method, we conduct a comparative analysis against eight state-of-the-art incomplete multi-view clustering methods. These methods are enumerated as follows:

\begin{itemize}
\item \textbf{DIMVC} \citep{xu2022deep}: This method proposes an imputation-free and fusion-free deep framework specifically designed for incomplete multi-view clustering.
\item \textbf{DSIMVC} \citep{tang2022deep}: This method introduces a bi-level optimization framework that addresses missing views by leveraging learned neighbor semantics.
\item \textbf{GIGA} \citep{yang2024geometric}: This method adaptively estimates the factual weight of each available view to mitigate the adverse effects of missing data.
\item \textbf{GIMVC} \citep{bai2024graph}: This is an imputation-free incomplete multi-view clustering method that incorporates a graph-guided mechanism.
\item \textbf{CDIMC-net} \citep{wen2021cdimc}: This method integrates encoders with a graph embedding strategy to capture both high-level features and local structural information.
\item \textbf{MRL\_CAL} \citep{wang2024contrastive2}: This method facilitates data recovery, consistent representation learning, and clustering through the joint learning of features across distinct subspaces.
\item \textbf{DCP} \citep{lin2022dual}: This method performs multi-view clustering by jointly maximizing inter-view mutual information and minimizing conditional entropy.

\item \textbf{GHICMC} \citep{chao2025global}: This method integrates representation learning, global graph propagation and contrastive clustering.
\end{itemize}

\subsection{Experimental Results and Analysis}
Table~\ref{tab:baselines} presents the average clustering results obtained over five runs for each method. It can be observed that our method consistently outperforms baseline methods on all datasets, under both complete and incomplete data conditions.

To further validate the effectiveness of our method, experiments are conducted with missing rates ranging from 0.1 to 0.7, with an interval of 0.2. Figure~\ref{fig:baselines} illustrates the performance of all methods across four datasets. It is evident that our proposed model achieves superior performance consistently across all datasets and missing rate settings.

\subsection{Ablation Study}
To validate the effectiveness of each component in our proposed method, we conducted ablation studies by progressively removing: (A) the masked graph reconstruction loss $L_{rec}$, (B) the embedding layer with the contrastive loss $L_{con}$, and (C) the self-supervised clustering module with $L_{kl}$. Table \ref{tab:ablation} presents the experimental results on the 100leaves dataset with a missing rate of 0.5. It is noteworthy that when the embedding layer is removed, view-specific graphs are constructed from the raw data $X$. The performance of the model consistently degrades to varying degrees upon the removal of each component. Notably, the absence of the embedding layer leads to a substantial decrease in performance. In contrast, the self-supervised clustering module demonstrates the least impact on the overall performance.

\begin{table}[ht]
\caption{Ablation studies on 100Leaves when $\delta=0.5$.}
\label{tab:ablation}
\centering
\begin{tabular}{cccccccccccccccc}
\toprule
\multicolumn{3}{c}{Components}&\multicolumn{3}{c}{100Leaves}\\
\cline{1-6}
A&B&C&ACC&NMI&ARI\\
\midrule
\checkmark&\checkmark&\checkmark&\textbf{0.812}&\textbf{0.891}&\textbf{0.703}\\
&\checkmark&\checkmark&0.704&0.824&0.561\\
\checkmark&&\checkmark&0.529&0.714&0.290\\
\checkmark&\checkmark&&0.806&0.890&0.698\\

\bottomrule
\end{tabular}
\end{table}

\subsection{Loss Function Comparison}
To validate the effectiveness of the masked graph reconstruction loss, we conduct experiments comparing its performance against the traditional graph reconstruction loss. Table \ref{tab:loss_comparison} summarizes the model performance on the Scene-15 dataset, evaluated under missing data rates ranging from 0.1 to 0.7, with an interval of 0.2. The results clearly demonstrate that the masked graph reconstruction loss consistently outperforms the traditional graph reconstruction loss across most experimental conditions.

\subsection{Parameter Sensitivity Analysis}
\begin{figure}[ht]
	\centering
	\includegraphics[width=0.5\textwidth]{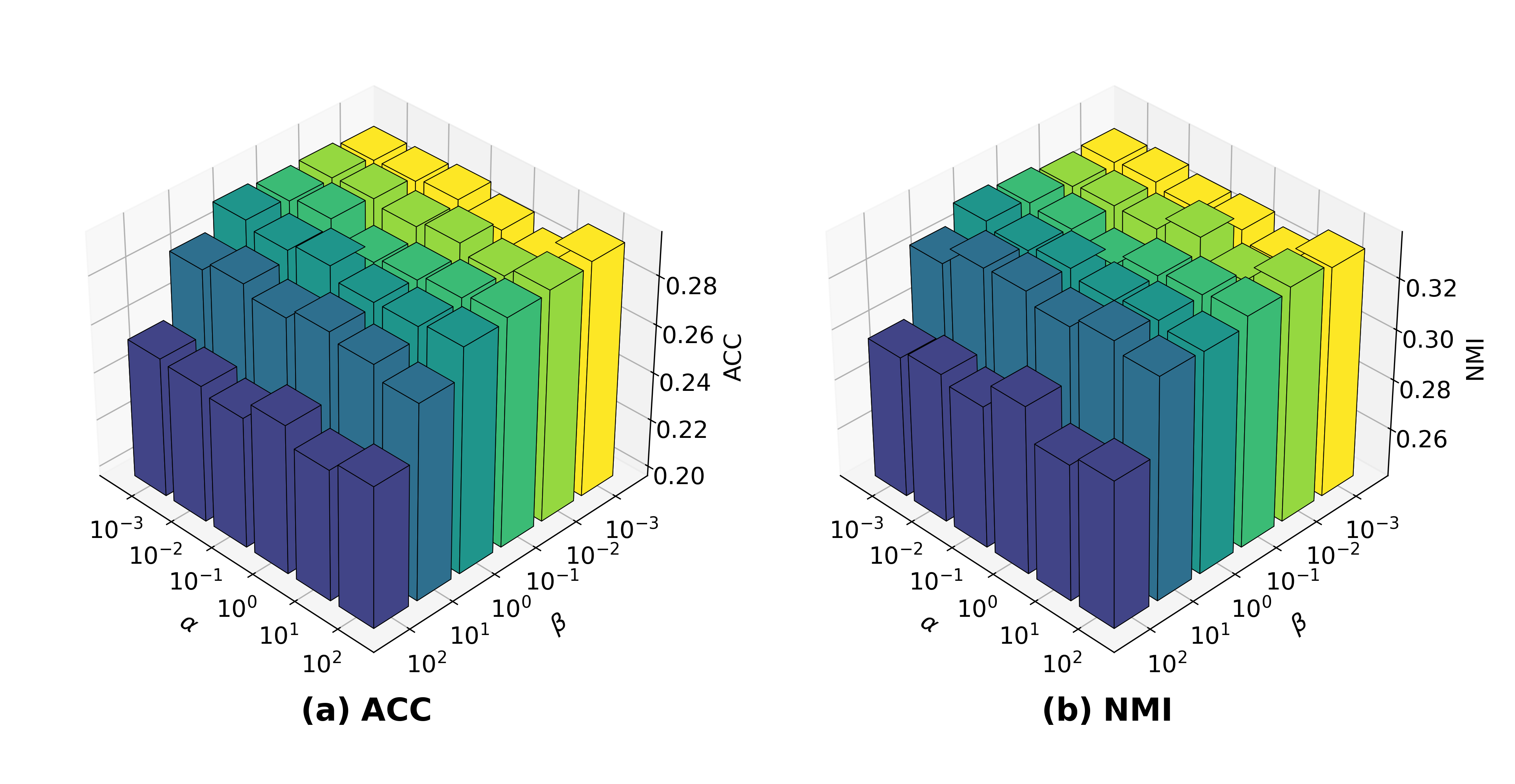} 
        \vspace{-7mm}
	\caption{Parameter sensitivity analysis for $\alpha$ and $\beta$ on Landuse-21 with missing rate of 0.5.}
	\label{fig:parameter_analysis}
\end{figure}

In this subsection, we analyze the model's sensitivity to hyperparameters $\alpha$ and $\beta$. Both hyperparameters are varied within the range $\{10^{-3}, 10^{-2}, 10^{-1}, 10^{0}, 10^{1}, 10^{2}\}$. Figure~\ref{fig:parameter_analysis} presents the experimental results obtained on the Landuse-21 dataset with a missing rate of 0.5. Our model demonstrates robustness to these hyperparameters, consistently achieving competitive performance across their specified reasonable ranges.

\begin{table}[ht]
\caption{Comparison of graph reconstruction loss functions on Scene-15 for varying $\delta$.}
\label{tab:loss_comparison}
\centering
\begin{tabular}{ccccccc}
\toprule
$\delta$ & Type of the loss & ACC & NMI & ARI \\
\midrule
\multirow{2}{*}{0.1} & masked        & \textbf{0.505}     & \textbf{0.497}     & \textbf{0.326}     \\
    & traditional   & 0.479     & 0.486     & 0.305     \\
\midrule
\multirow{2}{*}{0.3} & masked       & 0.469     & \textbf{0.464}     &  \textbf{0.298}    \\
    & traditional  & \textbf{0.470}     & 0.458     & 0.286     \\
\midrule
\multirow{2}{*}{0.5} & masked       & \textbf{0.452}     & \textbf{0.430}     & \textbf{0.273}     \\
    & traditional  & 0.447     & 0.425     &  0.262    \\
\midrule
\multirow{2}{*}{0.7} & masked       & \textbf{0.426}     & \textbf{0.397}     & \textbf{0.244}     \\
    & traditional  & 0.403     & 0.390     & 0.227     \\
\bottomrule
\end{tabular}
\end{table}

\section{Conclusion}
This paper proposes DGIMVCM, a novel incomplete multi-view clustering framework. The framework incorporates a GCN-based embedding layer, designed to dynamically extract view-specific graphs and effectively address the missing data problem. Furthermore, it employs a GAT-based encoder, which adaptively learns edge weights. To optimize this encoder and mitigate gradient noise during optimization, a masked graph reconstruction loss is utilized. Extensive experiments consistently demonstrate the superiority of DGIMVCM compared to state-of-the-art methods.

% \section{Acknowledgments}

\bibliography{aaai2026}

@article{fang2023comprehensive,
  title={A comprehensive survey on multi-view clustering},
  author={Fang, Uno and Li, Man and Li, Jianxin and Gao, Longxiang and Jia, Tao and Zhang, Yanchun},
  journal={IEEE Transactions on Knowledge and Data Engineering},
  volume={35},
  number={12},
  pages={12350--12368},
  year={2023},
  publisher={IEEE}
}

@inproceedings{chao2025global,
  title={Global Graph Propagation with Hierarchical Information Transfer for Incomplete Contrastive Multi-view Clustering},
  author={Chao, Guoqing and Xu, Kaixin and Xie, Xijiong and Chen, Yongyong},
  booktitle={Proceedings of the AAAI Conference on Artificial Intelligence},
  volume={39},
  number={15},
  pages={15713--15721},
  year={2025}
}

@inproceedings{chao2024incomplete,
  title={Incomplete contrastive multi-view clustering with high-confidence guiding},
  author={Chao, Guoqing and Jiang, Yi and Chu, Dianhui},
  booktitle={Proceedings of the AAAI Conference on Artificial Intelligence},
  volume={38},
  number={10},
  pages={11221--11229},
  year={2024}
}

@inproceedings{huang2023self,
  title={Self-supervised graph attention networks for deep weighted multi-view clustering},
  author={Huang, Zongmo and Ren, Yazhou and Pu, Xiaorong and Huang, Shudong and Xu, Zenglin and He, Lifang},
  booktitle={Proceedings of the AAAI Conference on Artificial Intelligence},
  volume={37},
  number={7},
  pages={7936--7943},
  year={2023}
}

@inproceedings{wang2024surer,
  title={Surer: Structure-adaptive unified graph neural network for multi-view clustering},
  author={Wang, Jing and Feng, Songhe and Lyu, Gengyu and Yuan, Jiazheng},
  booktitle={Proceedings of the AAAI Conference on Artificial Intelligence},
  volume={38},
  number={14},
  pages={15520--15527},
  year={2024}
}

@inproceedings{
chaofederated,
title={Federated Incomplete Multi-view Clustering with Globally Fused Graph Guidance},
author={Guoqing Chao and Zhenghao Zhang and Lei Meng and Jie Wen and Dianhui Chu},
booktitle={Forty-second International Conference on Machine Learning},
year={2025},
url={https://openreview.net/forum?id=7qvYLnJDRd}
}

@inproceedings{pu2024adaptive,
  title={Adaptive feature imputation with latent graph for deep incomplete multi-view clustering},
  author={Pu, Jingyu and Cui, Chenhang and Chen, Xinyue and Ren, Yazhou and Pu, Xiaorong and Hao, Zhifeng and Yu, Philip S and He, Lifang},
  booktitle={Proceedings of the AAAI Conference on Artificial Intelligence},
  volume={38},
  number={13},
  pages={14633--14641},
  year={2024}
}

@inproceedings{tang2022deep,
  title={Deep safe incomplete multi-view clustering: Theorem and algorithm},
  author={Tang, Huayi and Liu, Yong},
  booktitle={International Conference on Machine Learning},
  pages={21090--21110},
  year={2022},
  organization={PMLR}
}

@inproceedings{xu2022deep,
  title={Deep incomplete multi-view clustering via mining cluster complementarity},
  author={Xu, Jie and Li, Chao and Ren, Yazhou and Peng, Liang and Mo, Yujie and Shi, Xiaoshuang and Zhu, Xiaofeng},
  booktitle={Proceedings of the AAAI Conference on Artificial Intelligence},
  volume={36},
  number={8},
  pages={8761--8769},
  year={2022}
}

@inproceedings{xu2024deep,
  title={Deep variational incomplete multi-view clustering: Exploring shared clustering structures},
  author={Xu, Gehui and Wen, Jie and Liu, Chengliang and Hu, Bing and Liu, Yicheng and Fei, Lunke and Wang, Wei},
  booktitle={Proceedings of the AAAI Conference on Artificial Intelligence},
  volume={38},
  number={14},
  pages={16147--16155},
  year={2024}
}

@inproceedings{wang2024contrastive,
  title={Contrastive and view-interaction structure learning for multi-view clustering},
  author={Wang, Jing and Feng, Songhe},
  booktitle={Proceedings of the Thirty-Third International Joint Conference on Artificial Intelligence},
  pages={5055--5063},
  year={2024}
}

@inproceedings{li2015large,
  title={Large-scale multi-view spectral clustering via bipartite graph},
  author={Li, Yeqing and Nie, Feiping and Huang, Heng and Huang, Junzhou},
  booktitle={Proceedings of the AAAI Conference on Artificial Intelligence},
  volume={29},
  number={1},
  year={2015}
}

@article{mallah2013plant,
  title={Plant leaf classification using probabilistic integration of shape, texture and margin features},
  author={Mallah, Charles and Cope, James and Orwell, James and others},
  journal={Signal Processing, Pattern Recognition and Applications},
  volume={5},
  number={1},
  pages={45--54},
  year={2013}
}

@inproceedings{yang2010bag,
  title={Bag-of-visual-words and spatial extensions for land-use classification},
  author={Yang, Yi and Newsam, Shawn},
  booktitle={Proceedings of the 18th SIGSPATIAL International Conference on Advances in Geographic Information Systems},
  pages={270--279},
  year={2010}
}

@inproceedings{fei2005bayesian,
  title={A bayesian hierarchical model for learning natural scene categories},
  author={Fei-Fei, Li and Perona, Pietro},
  booktitle={2005 IEEE Computer Society Conference on Computer Vision and Pattern Recognition (CVPR'05)},
  volume={2},
  pages={524--531},
  year={2005},
  organization={IEEE}
}

@article{yang2024geometric,
  title={Geometric-inspired graph-based incomplete multi-view clustering},
  author={Yang, Zequn and Zhang, Han and Wei, Yake and Wang, Zheng and Nie, Feiping and Hu, Di},
  journal={Pattern Recognition},
  volume={147},
  pages={110082},
  year={2024},
  publisher={Elsevier}
}

@article{bai2024graph,
  title={Graph-guided imputation-free incomplete multi-view clustering},
  author={Bai, Shunshun and Zheng, Qinghai and Ren, Xiaojin and Zhu, Jihua},
  journal={Expert Systems with Applications},
  volume={258},
  pages={125165},
  year={2024},
  publisher={Elsevier}
}

@inproceedings{wen2021cdimc,
  title={CDIMC-net: cognitive deep incomplete multi-view clustering network},
  author={Wen, Jie and Zhang, Zheng and Xu, Yong and Zhang, Bob and Fei, Lunke and Xie, Guo-Sen},
  booktitle={Proceedings of the Twenty-Ninth International Conference on International Joint Conferences on Artificial Intelligence},
  pages={3230--3236},
  year={2021}
}

@article{wang2024contrastive2,
  title={Contrastive and adversarial regularized multi-level representation learning for incomplete multi-view clustering},
  author={Wang, Haiyue and Zhang, Wensheng and Ma, Xiaoke},
  journal={Neural Networks},
  volume={172},
  pages={106102},
  year={2024},
  publisher={Elsevier}
}

@article{vaswani2017attention,
  title={Attention is all you need},
  author={Vaswani, Ashish and Shazeer, Noam and Parmar, Niki and Uszkoreit, Jakob and Jones, Llion and Gomez, Aidan N and Kaiser, {\L}ukasz and Polosukhin, Illia},
  journal={Advances in Neural Information Processing Systems},
  volume={30},
  year={2017}
}

@article{lin2022dual,
  title={Dual contrastive prediction for incomplete multi-view representation learning},
  author={Lin, Yijie and Gou, Yuanbiao and Liu, Xiaotian and Bai, Jinfeng and Lv, Jiancheng and Peng, Xi},
  journal={IEEE Transactions on Pattern Analysis and Machine Intelligence},
  volume={45},
  number={4},
  pages={4447--4461},
  year={2022},
  publisher={IEEE}
}

@article{yang2018multi,
  title={Multi-view clustering: A survey},
  author={Yang, Yan and Wang, Hao},
  journal={Big Data Mining and Analytics},
  volume={1},
  number={2},
  pages={83--107},
  year={2018},
  publisher={TUP}
}

@article{zhou2024survey,
  title={A survey and an empirical evaluation of multi-view clustering approaches},
  author={Zhou, Lihua and Du, Guowang and Lue, Kevin and Wang, Lizheng and Du, Jingwei},
  journal={ACM Computing Surveys},
  volume={56},
  number={7},
  pages={1--38},
  year={2024},
  publisher={ACM New York, NY}
}

@article{yu_review_2024,
	title = {A review on multi-view learning},
	volume = {19},
	issn = {2095-2236},
	doi = {10.1007/s11704-024-40004-w},
	number = {7},
	journal = {Frontiers of Computer Science},
	author = {Yu, Zhiwen and Dong, Ziyang and Yu, Chenchen and Yang, Kaixiang and Fan, Ziwei and Chen, C. L. Philip},
	month = dec,
	year = {2024},
	pages = {197334},
}

@article{xu2013survey,
  title={A survey on multi-view learning},
  author={Xu, Chang and Tao, Dacheng and Xu, Chao},
  journal={arXiv preprint arXiv:1304.5634},
  year={2013}
}

@article{fu2020overview,
  title={An overview of recent multi-view clustering},
  author={Fu, Lele and Lin, Pengfei and Vasilakos, Athanasios V and Wang, Shiping},
  journal={Neurocomputing},
  volume={402},
  pages={148--161},
  year={2020},
  publisher={Elsevier}
}

@article{tang2024incomplete,
  title={Incomplete multi-view learning: Review, analysis, and prospects},
  author={Tang, Jingjing and Yi, Qingqing and Fu, Saiji and Tian, Yingjie},
  journal={Applied Soft Computing},
  volume={153},
  pages={111278},
  year={2024},
  publisher={Elsevier}
}

@inproceedings{cheng2021multi,
  title={Multi-view attribute graph convolution networks for clustering},
  author={Cheng, Jiafeng and Wang, Qianqian and Tao, Zhiqiang and Xie, Deyan and Gao, Quanxue},
  booktitle={Proceedings of the Twenty-ninth International Conference on International Joint Conferences on Artificial Intelligence},
  pages={2973--2979},
  year={2021}
}

@inproceedings{wen2021structural,
  title={Structural deep incomplete multi-view clustering network},
  author={Wen, Jie and Wu, Zhihao and Zhang, Zheng and Fei, Lunke and Zhang, Bob and Xu, Yong},
  booktitle={Proceedings of the 30th ACM international conference on information \& knowledge management},
  pages={3538--3542},
  year={2021}
}

@article{shao2022heterogeneous,
  title={Heterogeneous graph neural network with multi-view representation learning},
  author={Shao, Zezhi and Xu, Yongjun and Wei, Wei and Wang, Fei and Zhang, Zhao and Zhu, Feida},
  journal={IEEE Transactions on Knowledge and Data Engineering},
  volume={35},
  number={11},
  pages={11476--11488},
  year={2022},
  publisher={IEEE}
}

@inproceedings{wen2024dual,
  title={Dual-Optimized Adaptive Graph Reconstruction for Multi-View Graph Clustering},
  author={Wen, Zichen and Wu, Tianyi and Ren, Yazhou and Ling, Yawen and Cui, Chenhang and Pu, Xiaorong and He, Lifang},
  booktitle={Proceedings of the 32nd ACM International Conference on Multimedia},
  pages={1819--1828},
  year={2024}
}

@inproceedings{lazebnik2006beyond,
  title={Beyond bags of features: Spatial pyramid matching for recognizing natural scene categories},
  author={Lazebnik, Svetlana and Schmid, Cordelia and Ponce, Jean},
  booktitle={2006 IEEE Computer Society Conference on Computer Vision and Pattern Recognition (CVPR'06)},
  volume={2},
  pages={2169--2178},
  year={2006},
  organization={IEEE}
}

@inproceedings{
renmulti,
title={Multi-View Graph Clustering via Node-Guided Contrastive Encoding},
author={Yazhou Ren and Junlong Ke and Zichen Wen and Tianyi Wu and Yang Yang and Xiaorong Pu and Lifang He},
booktitle={Forty-second International Conference on Machine Learning},
year={2025},
url={https://openreview.net/forum?id=Ae5qnQxAxQ}
}

@inproceedings{ren2024dynamic,
  title={Dynamic weighted graph fusion for deep multi-view clustering},
  author={Ren, Yazhou and Pu, Jingyu and Cui, Chenhang and Zheng, Yan and Chen, Xinyue and Pu, Xiaorong and He, Lifang},
  booktitle={Proceedings of the Thirty-Third International Joint Conference on Artificial Intelligence},
  pages={4842--4850},
  year={2024}
}

@article{xu2021deep,
  title={Deep embedded multi-view clustering with collaborative training},
  author={Xu, Jie and Ren, Yazhou and Li, Guofeng and Pan, Lili and Zhu, Ce and Xu, Zenglin},
  journal={Information Sciences},
  volume={573},
  pages={279--290},
  year={2021},
  publisher={Elsevier}
}

@article{wu2024self,
  title={Self-weighted contrastive fusion for deep multi-view clustering},
  author={Wu, Song and Zheng, Yan and Ren, Yazhou and He, Jing and Pu, Xiaorong and Huang, Shudong and Hao, Zhifeng and He, Lifang},
  journal={IEEE Transactions on Multimedia},
  volume={26},
  pages={9150--9162},
  year={2024},
  publisher={IEEE}
}

@article{yu2025incomplete,
  title={Incomplete Multi-View Clustering via Mutual Information},
  author={Yu, Xuejiao and Chao, Guoqing and Jiang, Yi and Ke, Guanzhou and Chu, Dianhui},
  journal={IEEE Transactions on Multimedia},
  year={2025},
  publisher={IEEE}
}

\newpage
\appendix
\onecolumn

\section{A\quad Derivation of gradient analysis for graph reconstruction loss}

\subsection{Traditional graph reconstruction loss}
Without loss of generality, consider the $i$-th sample in view $v$. The traditional graph reconstruction loss is defined as:
\begin{align}
    \label{aeq:loss_rec_t}
   L_{rec\_t,i}^v=\sum_{j=1}^N(\hat{A}^v_{ij}-\overline{A}_{ij})^2.
\end{align}
The gradient of this loss with respect to feature $h_i^v$ is:
\begin{align}
    \label{aeq:loss_rec_t_gra1}
   \frac{\partial L^v_{rec\_t,i}}{\partial h^v_i}=2\sum_{j=1}^N(\hat{A}^v_{ij}-\overline{A}_{ij})\frac{\partial \hat{A}^v_{ij}}{\partial h^v_i}.
\end{align}
For the reconstructed graph calculated by the radial basis function: $\hat{A}^v_{ij}=e^{-\frac{||h_i^v-h_j^v||^2}{t}}$, the gradient of $\hat{A}^v_{ij}$ with respect to $h^v_i$ is:
\begin{equation}
\label{aeq:A_gra}
   \frac{\partial \hat{A}^v_{ij}}{\partial h^v_i}=\frac{2}{t}\hat{A}^v_{ij}(h^v_j-h^v_i).
\end{equation}
Substituting Eq.~\eqref{aeq:A_gra} into Eq.~\eqref{aeq:loss_rec_t_gra1}, we obtain:
\begin{equation}
\label{aeq:loss_rec_t_gra2}
   \frac{\partial L^v_{rec\_t,i}}{\partial h^v_i}=\frac{4}{t}\sum_{j=1}^N(\hat{A}^v_{ij}-\overline{A}_{ij})\hat{A}^v_{ij}(h^v_j-h^v_i).
\end{equation}
Let $E_i=\{j \mid \overline{A}_{ij}=1\}$ denote the set of indices of samples adjacent to sample $i$ in the global graph. The gradient $\frac{\partial L^v_{rec\_t,i}}{\partial h^v_i}$ can be decomposed into two parts:
\begin{align}
   \frac{\partial L^v_{rec\_t,i}}{\partial h^v_i}=\frac{4}{t}\left(\sum_{j\in E_i}(\hat{A}^v_{ij}-1)\hat{A}^v_{ij}(h^v_j-h^v_i)\right.\left.+\sum_{j\notin E_i}{(\hat{A}^v_{ij}})^2(h^v_j-h_i^v)\right).
   \label{aeq:loss_rec_t_gra}
\end{align}

\subsection{Masked graph reconstruction loss}
For the $i$-th sample in view $v$, our masked graph reconstruction loss is defined as:
\begin{align}
    L_{rec\_m,i}^v=\sum_{j=1}^N(M^v_{ij}\hat{A}^v_{ij}-\overline{A}_{ij})^2.
\end{align}
The gradient of $L_{rec\_m,i}^v$ with respect to $h^v_i$ is:
\begin{align}
    \frac{\partial L^v_{rec\_m,t}}{\partial h^v_i}&=2\sum_{j=1}^N(M_{ij}^v\hat{A}_{ij}^v-\overline{A}_{ij})M_{ij}^v\frac{\partial \hat{A}^v_{ij}}{\partial h^v_i}\\
    &=\frac{4}{t}\sum_{j=1}^N(M_{ij}^v\hat{A}_{ij}^v-\overline{A}_{ij})M^v_{ij}\hat{A}^v_{ij}(h^v_j-h^v_i)\notag\\
    &=\frac{4}{t}\sum_{j:M_{ij}^v=1}(\hat{A}^v_{ij}-\overline{A}_{ij})\hat{A}^v_{ij}(h^v_j-h^v_i)\notag\\
    &=\frac{4}{t}\left(\sum_{j:j\in E_i,M^v_{ij}=1}(\hat{A_{ij}^v}-1)\hat{A}^v_{ij}(h^v_j-h^v_i) \right.\left.+\sum_{j:j\notin E_i,M^v_{ij}=1}{\hat{A_{ij}^v}}^2(h^v_j-h^v_i)\right).
\end{align}

\section{B\quad Datasets}
We conduct evaluations on four widely used public datasets: Handwritten (HW), Scene-15, Landuse-21, and 100leaves. Detailed information regarding these datasets is presented in Table~\ref{tab:datasets}.
\begin{table}[ht]
\caption{Description of the datasets.}
\label{tab:datasets}
\centering
\begin{tabular}{c|c|c|c|c}
\toprule
Datasets & Samples & Views & Dimensions & Classes \\ 
\midrule
Scene-15 & 4485   & 3 & 20/59/40 & 15 \\
HW & 2000  & 6 & 240/76/216/47/64/6 & 10 \\
Landuse-21 & 2100 & 3   & 20/59/40  & 21 \\
100leaves & 1600 & 3   & 64/64/64 & 100 \\
\bottomrule
\end{tabular}
\end{table}

\section{C\quad Detailed experimental results}
\subsection{C.1\quad Experimental setup}
DGIMVCM is implemented using PyTorch 2.7.1 on Ubuntu 20.04. The Adam optimizer is employed to minimize the loss function. The scale parameter $t$ is set to 2, and the temperature parameter $\tau$ is set to 0.5. The hyperparameters $\alpha$ and $\beta$ within the loss function are both set to 1.

\subsection{C.2\quad More Experiments of DGIMVCM} 
\begin{figure*}[ht]
	\centering
	\includegraphics[width=\textwidth]{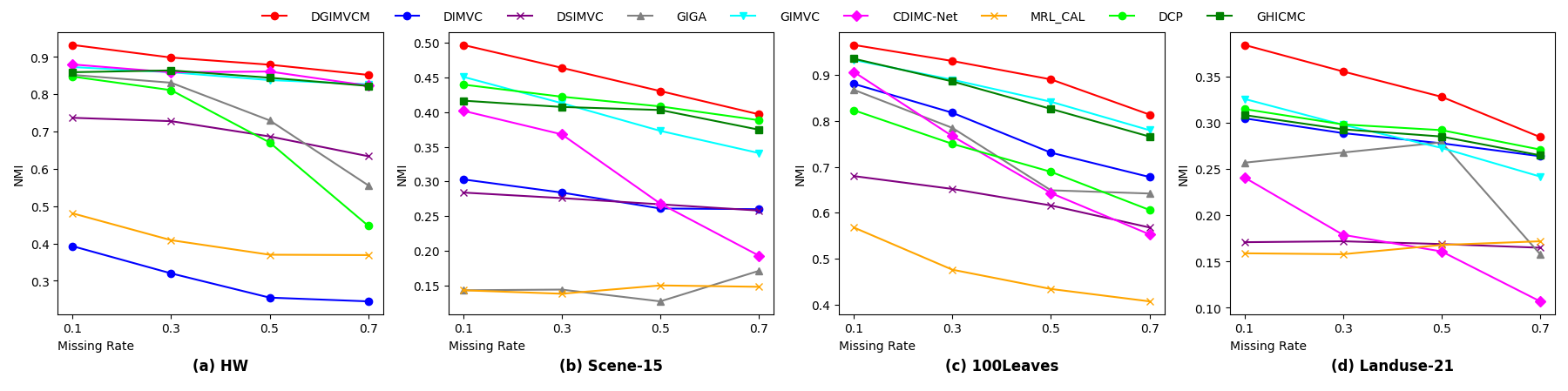}
        \vspace{-8mm}
	\caption{NMI on four datasets with different missing rates. (a) HW, (b) Scene-15, (c) 100Leaves, (d) Landuse-21.}
	\label{fig:NMI_baselines}
\end{figure*}

\begin{figure*}[ht]
	\centering
	\includegraphics[width=\textwidth]{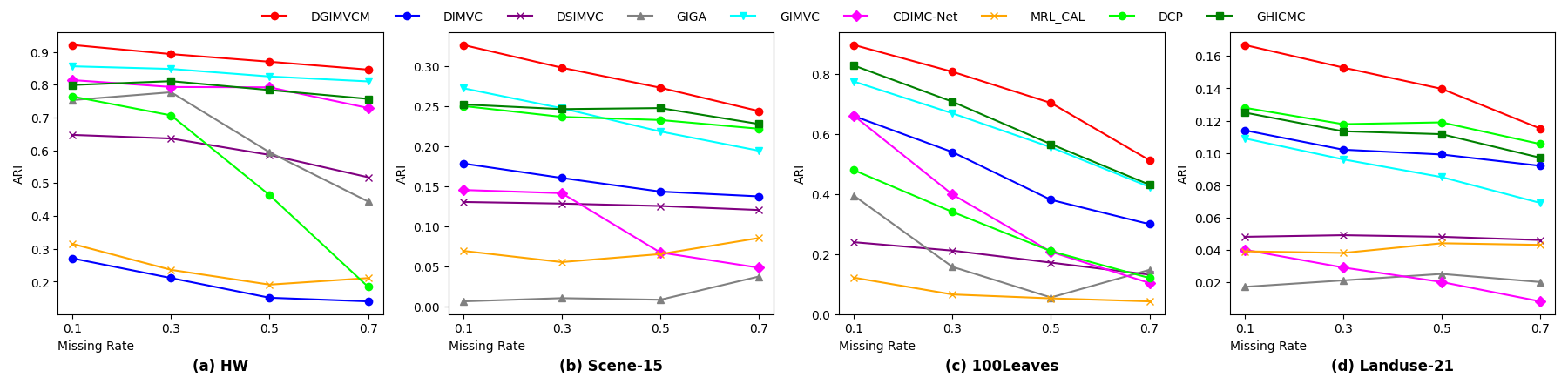}
        \vspace{-8mm}
	\caption{ARI on four datasets with different missing rates. (a) HW, (b) Scene-15, (c) 100Leaves, (d) Landuse-21.}
	\label{fig:ARI_baselines}
\end{figure*}

We present the comparative experimental results, evaluated using NMI and ARI, for all methods across various missing rates. The missing rates range from 0.1 to 0.7, with an interval of 0.2. The NMI and ARI results are depicted in Figure~\ref{fig:NMI_baselines} and Figure~\ref{fig:ARI_baselines}, respectively.

\subsection{C.3\quad More Ablation Study}
This section presents further ablation studies on the Scene-15 and HW datasets, with results summarized in Table \ref{tab:ablation_more}. The components in Table~\ref{tab:ablation_more} are defined as follows: (A) the masked graph reconstruction loss $L_{rec}$, (B) the embedding layer with the contrastive loss $L_{con}$, and (C) the self-supervised clustering module with $L_{kl}$.

\begin{table}[ht]
\caption{Ablation study on Scene-15 and HW when $\delta=0.5$.}
\label{tab:ablation_more}
\centering
\begin{tabular}{ccccccccccccccc}
\toprule
\multicolumn{3}{c}{Components}&\multicolumn{3}{c}{Scene-15}&\multicolumn{3}{c}{HW}\\
\cline{1-9}
A&B&C&ACC&NMI&ARI&ACC&NMI&ARI\\
\midrule
\checkmark&\checkmark&\checkmark&\textbf{0.452}&\textbf{0.430}&\textbf{0.273}&\textbf{0.939}&\textbf{0.879}&\textbf{0.870}\\
&\checkmark&\checkmark&0.322&0.316&0.158&0.809&0.774&0.689\\
\checkmark&&\checkmark&0.246&0.208&0.093&0.547&0.570&0.404\\
\checkmark&\checkmark&&0.446&0.430&0.272&0.936&0.877&0.865\\

\bottomrule
\end{tabular}
\end{table}

\subsection{C.4\quad Further Loss Function Comparison}
We further evaluate the performance of masked reconstruction loss against traditional reconstruction loss on the Landuse-21 dataset under various missing rates. The empirical results, summarized in Table \ref{tab:loss_comparison_landuse}, demonstrate that the proposed masked reconstruction loss consistently achieves superior performance.

\begin{table}[ht]
\caption{Comparison of graph reconstruction loss functions on Landuse-21 for varying $\delta$.}
\label{tab:loss_comparison_landuse}
\centering
\begin{tabular}{ccccccc}
\toprule
$\delta$ & Type of the loss & ACC & NMI & ARI \\
\midrule
\multirow{2}{*}{0.1} & masked        & \textbf{0.313}     & \textbf{0.384}     & \textbf{0.167}     \\
    & traditional   & 0.302     & 0.373     & 0.158     \\
\midrule
\multirow{2}{*}{0.3} & masked       & \textbf{0.303}      & \textbf{0.356 }     &  \textbf{0.153 }    \\
    & traditional  & 0.288     &  0.347     &  0.146     \\
\midrule
\multirow{2}{*}{0.5} & masked       & \textbf{0.287 }     & \textbf{0.328 }     & \textbf{0.140 }     \\
    & traditional  &  0.281     & 0.320      &  0.134     \\
\midrule
\multirow{2}{*}{0.7} & masked       & \textbf{0.264 }     & \textbf{0.285 }     & \textbf{0.115 }     \\
    & traditional  & 0.252      & 0.272      & 0.108      \\
\bottomrule
\end{tabular}
\end{table}

\subsection{C.5\quad Further Parameter Sensitivity Analysis}
\begin{figure}[ht]
	\centering
	\includegraphics[width=0.3\textwidth]{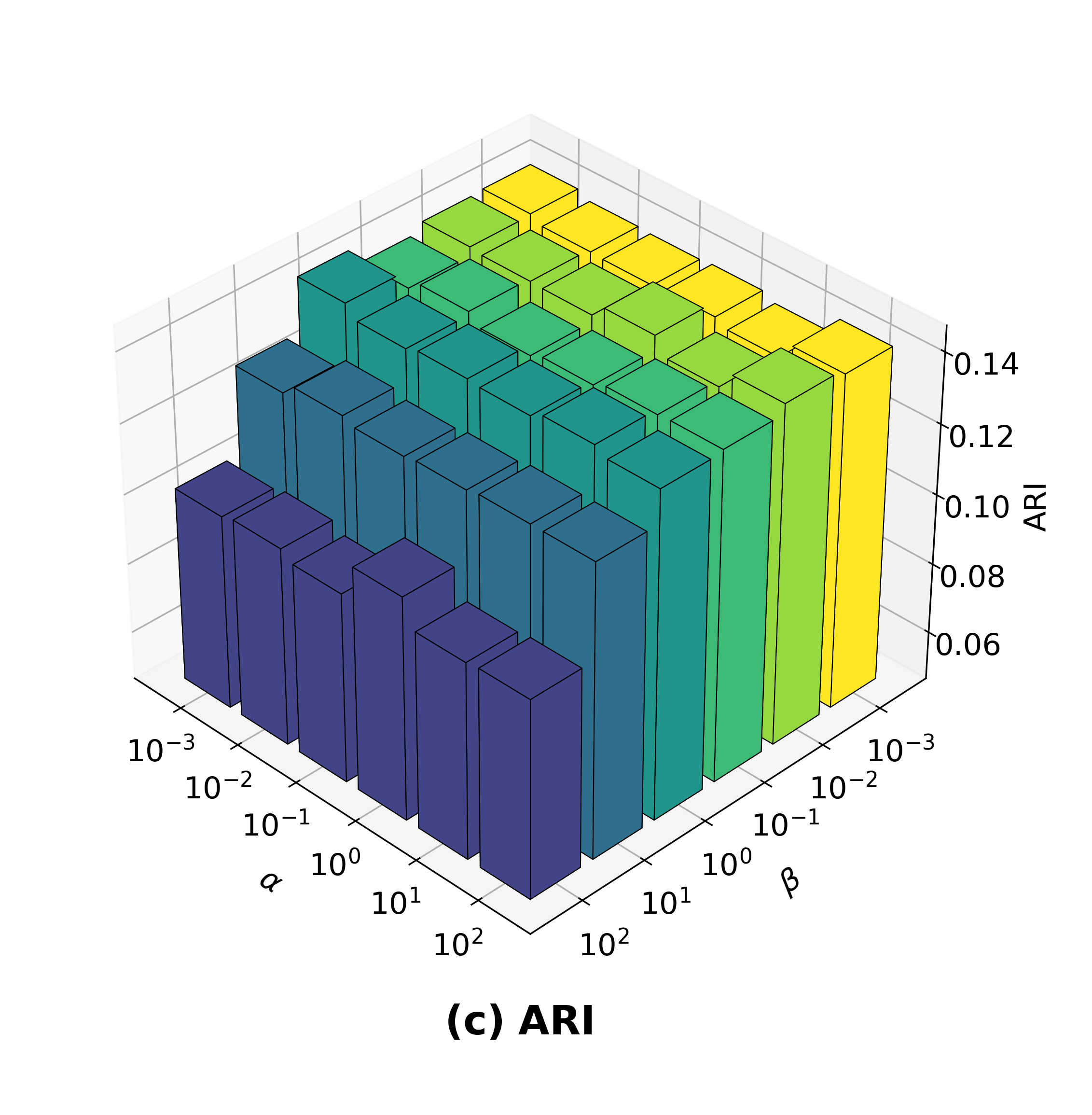} 
        \vspace{-7mm}
	\caption{Parameter sensitivity analysis for $\alpha$ and $\beta$ on Landuse-21 with missing rate of 0.5 using ARI as metric.}
	\label{fig:parameter_analysis_ari_landuse}
\end{figure}

Due to space contraints, the main body of this paper only presents the sensitivity analysis for parameters $\alpha$ and $\beta$ on the Landuse-21 dataset at a missing rate of 0.5 by using ACC and NMI as metric. Here, we additionally illustrate the ARI performance in Figure~\ref{fig:parameter_analysis_ari_landuse}. 

\begin{figure}[ht]
	\centering
	\includegraphics[width=0.8\textwidth]{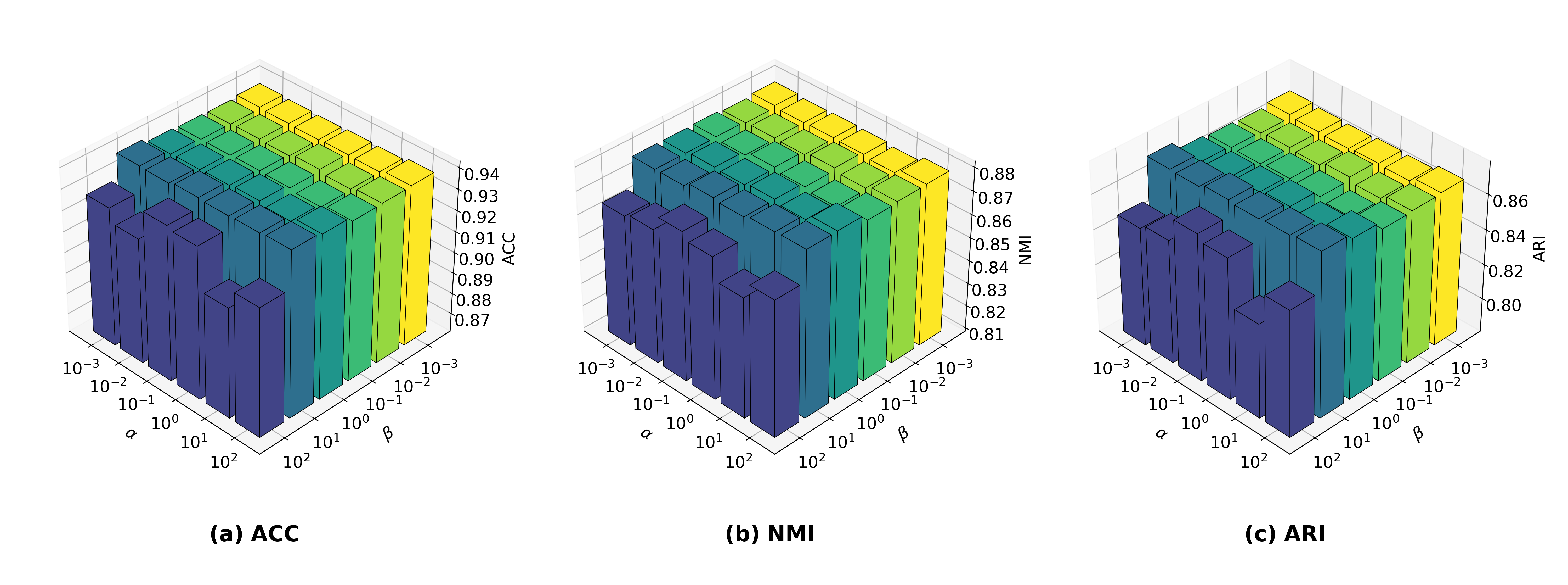} 
        \vspace{-7mm}
	\caption{Parameter sensitivity analysis for $\alpha$ and $\beta$ on HW with missing rate of 0.5.}
	\label{fig:parameter_analysis_HW}
\end{figure}

Furthermore, we conducted a parameter sensitivity analysis on the HW dataset with a missing rate of 0.5, and the results are depicted in Figure~\ref{fig:parameter_analysis_HW}. It is evident from these analyses that our model exhibits robustness to variations in $\alpha$ and $\beta$, consistently achieving strong performance across a reasonable range of parameter values.

\subsubsection{C.6\quad Visualization Analysis}
\begin{figure}[htbp] 
    \centering % 
 
    \begin{subfigure}[b]{0.23\textwidth} 
        \centering
        \includegraphics[width=\textwidth]{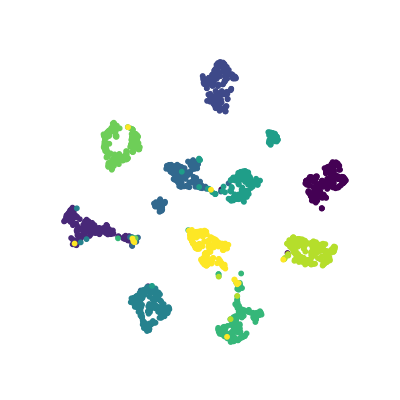} % 插入第一个子图，width=\textwidth表示图片填充子图的宽度
        \caption{missing rate of 0.1} % 第一个子图的标题
        \label{fig:vis_ms_0_1} % 子图 A 的引用标签
    \end{subfigure}
    \hfill % 在子图之间插入水平空间，使其均匀分布
    \begin{subfigure}[b]{0.23\textwidth}
        \centering
        \includegraphics[width=\textwidth]{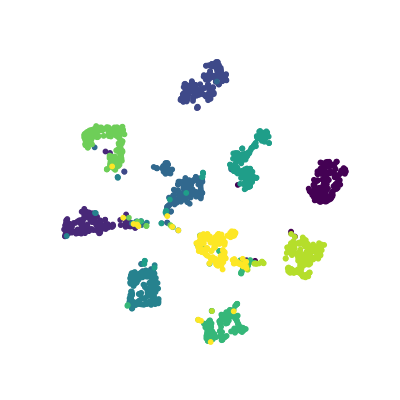}
        \caption{missing rate of 0.3}
        \label{fig:vis_ms_0_3}
    \end{subfigure}
    \hfill
    % 第二行
    \begin{subfigure}[b]{0.23\textwidth}
        \centering
        \includegraphics[width=\textwidth]{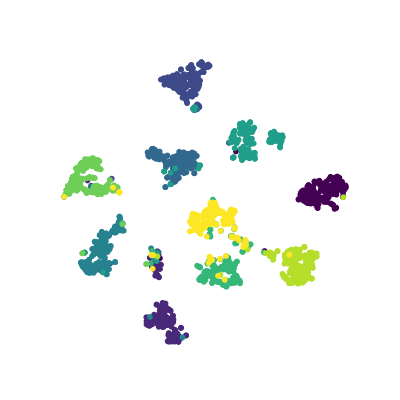}
        \caption{missing rate of 0.5}
        \label{fig:vis_ms_0_5}
    \end{subfigure}
    \hfill
    \begin{subfigure}[b]{0.23\textwidth}
        \centering
        \includegraphics[width=\textwidth]{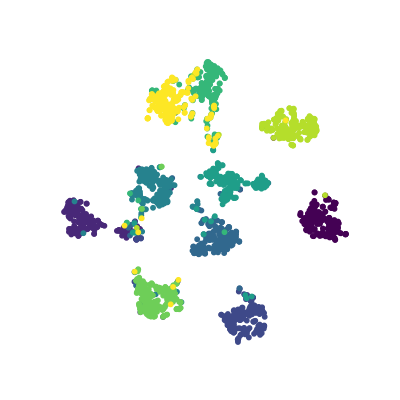} % LaTeX自带的一个通用示例图
        \caption{missing rate of 0.7}
        \label{fig:vis_ms_0_7}
    \end{subfigure}
 
    \caption{The visualization results on HandWritten dataset
 with different missing rates.} % 整个图的总标题
    \label{fig:visualization} % 整个图的引用标签
\end{figure}

We visualize the clustering results on the HandWritten dataset with different missing rates using t-SNE, as shown in Figure~\ref{fig:visualization}. It can be observed that our model effectively separates features from different classes. Furthermore, we note that as the missing rate increases, although the clustering performance degrades compared to lower missing rates, a clear cluster structure is still maintained.

\end{document}